\pdfoutput=1

\documentclass[11pt]{article}

\usepackage[final]{acl}

\usepackage{times}
\usepackage{latexsym}
\usepackage{array}
\usepackage{booktabs}
\usepackage{multirow}
\usepackage{amsmath} 
\usepackage{algorithm}
\usepackage{algpseudocode}

\usepackage{pifont}
\usepackage{amssymb}

\usepackage[T1]{fontenc}

\usepackage[utf8]{inputenc}

\usepackage{microtype}

\usepackage{graphicx}
\usepackage{enumitem}
\usepackage{xcolor}
\usepackage{tcolorbox}
\usepackage{colortbl}
\usepackage{makecell}
\usepackage[normalem]{ulem} 
\usepackage{tcolorbox}
\definecolor{aliceblue}{RGB}{255, 238, 241}

\definecolor{babyred}{rgb}{0.85, 0.93, 0.97}

\definecolor{babygreen}{rgb}{0.85, 0.97, 0.85}

\definecolor{uclablue}{RGB}{159, 195, 224}

\definecolor{uclagold}{RGB}{255, 240, 180}

\definecolor{intro_green}{RGB}{226,240,217}
\definecolor{intro_blue}{RGB}{221,235,247}
\definecolor{mygray}{gray}{.9}

\definecolor{my_green}{RGB}{51,102,0}
\definecolor{my_red}{RGB}{204, 0, 0}

\title{SynWorld: Virtual Scenario Synthesis for Agentic Action\\ Knowledge Refinement}

\author{
  Runnan Fang$^{\spadesuit}$, 
  Xiaobin Wang$^{\heartsuit}$,
  Yuan Liang$^{\spadesuit}$, 
  \textbf{Shuofei Qiao}$^{\spadesuit}$,  
  \textbf{Jialong Wu}$^{\heartsuit}$,  
  \textbf{Zekun Xi}$^{\spadesuit}$,\\
  \textbf{Ningyu Zhang}$^{\spadesuit}$\footnotemark[1]~, 
  \textbf{Yong Jiang}$^{\heartsuit}$\footnotemark[1]~, 
  \textbf{Pengjun Xie}$^{\heartsuit}$,
  \textbf{Fei Huang}$^{\heartsuit}$,
  \textbf{Huajun Chen}$^{\spadesuit}$ $^\clubsuit$\thanks{$\quad$ Corresponding Author.}\\
  $^\spadesuit$Zhejiang University ~
  $^\heartsuit$Alibaba Group\\
 $^\clubsuit$Zhejiang Key Laboratory of Big Data Intelligent Computing \\
  \texttt{\{rolnan,zhangningyu\}@zju.edu.cn} \\
}

\begin{document}
\maketitle
\begin{abstract}
In the interaction between agents and their environments, agents expand their capabilities by planning and executing actions. However, LLM-based agents face substantial challenges when deployed in novel environments or required to navigate unconventional action spaces. To empower agents to autonomously explore environments, optimize workflows, and enhance their understanding of actions, we propose SynWorld, a framework that allows agents to synthesize possible scenarios with multi-step action invocation within the action space and perform Monte Carlo Tree Search (MCTS) exploration to effectively refine their action knowledge in the current environment. Our experiments demonstrate that SynWorld is an effective and general approach to learning action knowledge in new environments\footnote{Code is available at \url{https://github.com/zjunlp/SynWorld}.}.
\end{abstract}

\section{Introduction}
By leveraging decision-making capabilities to execute task-oriented actions within dynamic environments, Large Language Models (LLM) based agents demonstrate enhanced environmental interactivity and operational versatility~\cite{DBLP:conf/iccv/SongSWCW023,DBLP:conf/iclr/0036YZXLL0DMYZ024,wu2025webwalker,xi2025omnithinkexpandingknowledgeboundaries,shi2025towards,qu2025survey}.
In the real world, agents perform actions by leveraging tools like web search engines~\cite{DBLP:conf/kdd/FanDNWLYCL24,zhao2024retrieval,ning2025survey} or API calls \cite{DBLP:conf/eccv/LiuCLZLRZYSZZGL24,tao2024harnessing} to access feedback from the real world, which addresses the static knowledge limitations of LLMs, facilitating a deeper comprehension of the real world. 
It is crucial for agents to learn how to plan and execute actions in the environment.
Nonetheless, as the complexity of tasks increases and novel environments emerge, manually annotated environment descriptions and a predefined action documents~\cite{DBLP:journals/corr/abs-2410-08197,sri2024automating,zhang2025igniting} for agents are often not consistent with the actual environmental conditions and action usage \cite{DBLP:journals/corr/abs-2407-06249,huang2024towards}. 
Refining well-defined and aligned descriptions of the environment and actions is time-consuming and labor-intensive.
\begin{figure}[tb]
    \centering
    \includegraphics[width=1\columnwidth]{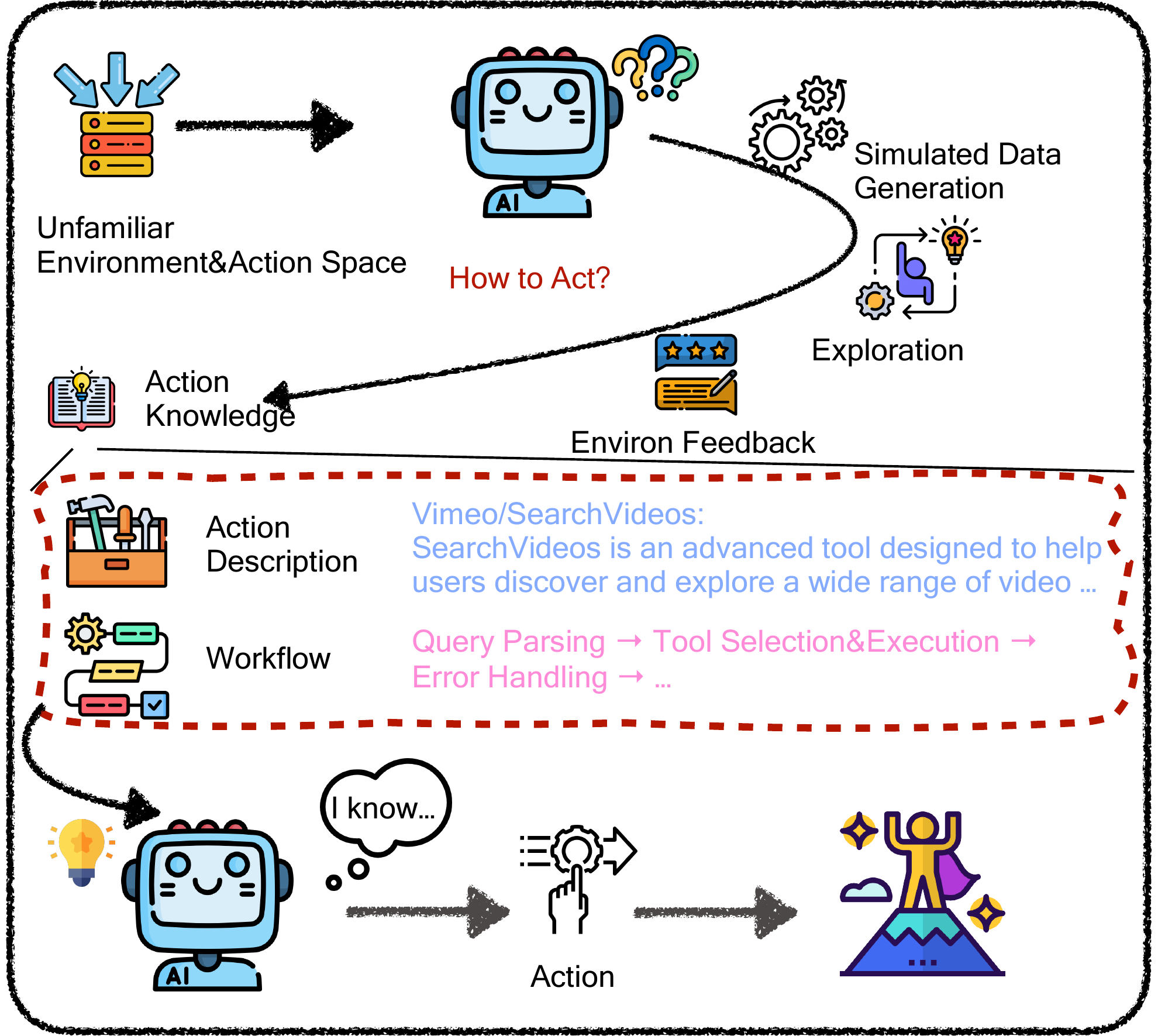}
    \caption{
    Our method with exploration to refine action knowledge in Synthesized Scenario.}
    \label{fig:overview}
    \vspace{-3mm}
\end{figure}

Therefore, to master inexperienced action and complicated task requirements in new complex environments, refinement of the agentic action knowledge is essential.
Previous studies have explored the acquisition of action knowledge through feedback in scenarios synthesized by LLMs.
Similar to the way humans acquire skills through trial and error, agents can also optimize the descriptions of actions by leveraging feedback from simulated scenarios~\cite{DBLP:journals/corr/abs-2401-06201,du2024anytool,bouzenia2024repairagent}.
However, these methods exhibit two critical limitations: 
(1) The synthetic scenarios they utilize are often restricted to single-action, which hinders agents from learning workflows suitable for these tasks, and
(2) The linear iterative optimization process lacks a clear direction for improvement, making it susceptible to stagnation and quickly reaching its performance ceiling.

To address these limitations, we propose a new framework, \textbf{SynWorld}, designed to assist agents in learning unfamiliar actions in new environments as shown in Figure~\ref{fig:overview}.
\textbf{SynWorld} first synthesizes virtual scenarios involving multiple coordinated actions.
Then through iterative MCTS optimization in the exploration of virtual scenarios, the framework enables more thorough and bidirectional refinement between action descriptions and workflow patterns, ensuring better alignment with environmental constraints.
Experiments demonstrate that action knowledge can be learned in virtual environments and effectively generalized to the real world, with optimization through MCTS exploration.

\section{Background}


\subsection{Agent Planning}

An agent interacts with its environment by perceiving its state, selecting actions to achieve a goal, and learning from feedback in the form of rewards.
Its framework consists of a state space \(\mathcal{S}\) that represents the environment's properties, an action space \(\mathcal{A}\) that defines allowable interactions, and an observation space \(\Omega\) for perceptual inputs. Progress toward task \(T\) is measured through a reward function \(\mathcal{R}\). Central to decision-making is a planning mechanism \(\mathcal{P}_\theta\), where \(\pi_\theta\) are fixed model weights. The agent's architecture is defined by the tuple:  
\begin{align}
  \mathcal{P}_\theta = \pi_\theta(\mathcal{S}, \mathcal{A}, \Omega, \mathcal{R})  
\end{align}
 This formula delineates the manner in which an agent assesses its current state and interprets environmental feedback to generate plans.

\subsection{Action Knowledge}

Action knowledge \(\mathcal{AK}\) serves as the strategic foundation governing an agent’s adaptive behavior in dynamic and unfamiliar environments. It contains \textbf{action description} about the awareness of executable actions with \textbf{cognitive workflows} about task decomposition and action sequences.

\section{Method}

\begin{figure*}[tb]
    \centering
    \includegraphics[width=\linewidth]{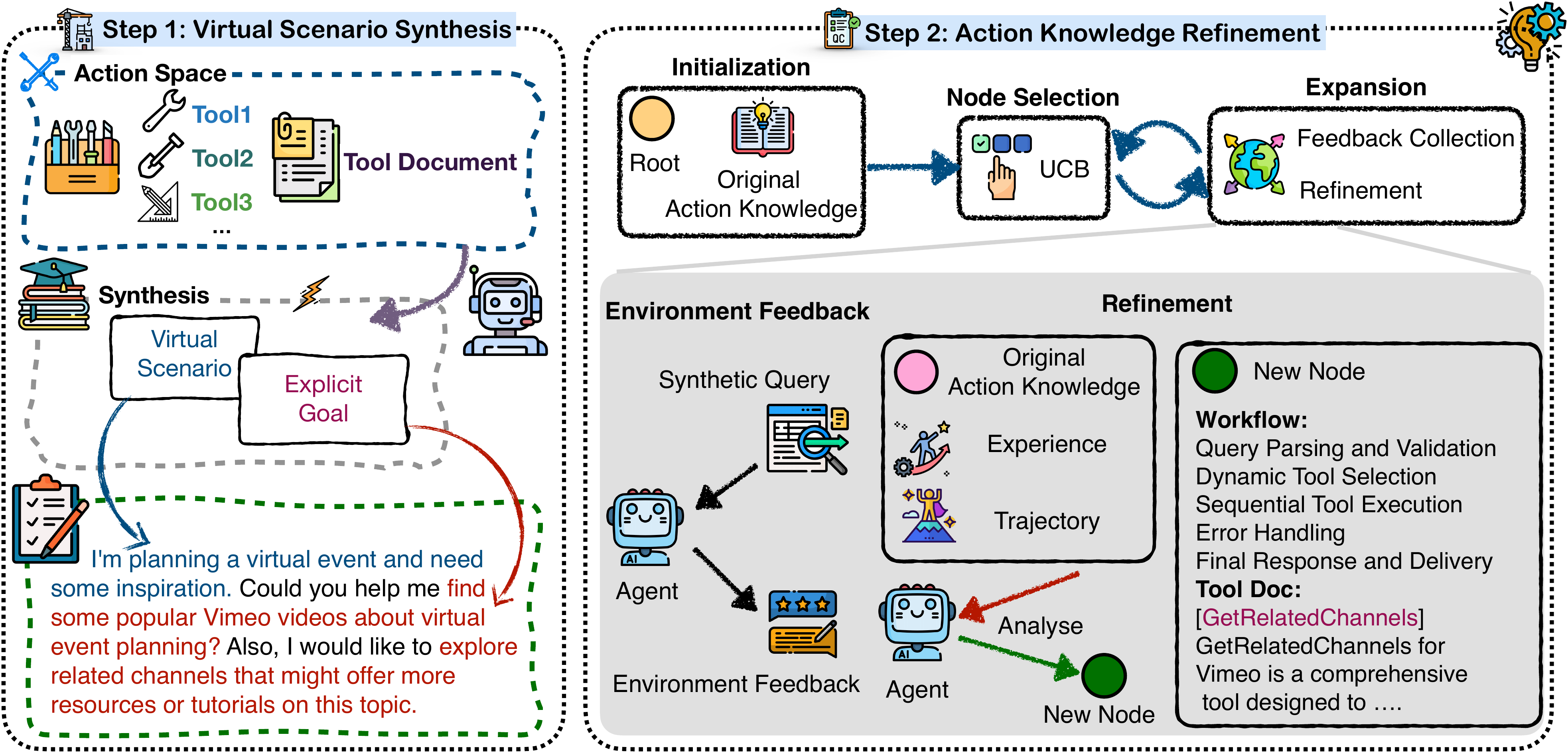}
    \caption{\textbf{The overall framework of SynWorld}: we first extract composable tools from the toolkit to generate new scenes and tasks. Then, we allow agents to explore the synthesized virtual scenes using MCTS to optimize action knowledge, thereby learning how to execute actions and plan tasks.}
    \label{fig:exper3}
    \vspace{-2mm}
\end{figure*}

In this section, we begin by detailing how to utilize the action space to synthesize scenarios and specific objectives. Subsequently, we dive into the application of MCTS to explore and discover action knowledge within these synthesized scenarios. 
The SynWorld framework is shown in Figure~\ref{fig:exper3}.
\subsection{Scenario Synthesis}
To address generalization challenges in multistep tool operationalization, we propose a framework that synthesizes scenarios through tool-conditioned task generation. Our methodology formalizes scenario synthesis as:  
\begin{align}
\mathcal{S}(t) = \{( \mathcal{B}, \mathcal{G} ) \mid \forall t \subseteq T)\},
\end{align}
where a subset of tools $t$ selected by llm from the complete set of tools $T$ to design a scenario. Each scenario comprises two part:
\textbf{Background} $\mathcal{B}$: The contextual scenario specifying initial conditions and constraints;
\textbf{Goal} $\mathcal{G}$: The terminal objective requiring tool-mediated resolution. We provide examples using a few-shot approach to enable the llm to synthesize queries.

The mapping enforces that distinct tool combinations yield nontrivial scenario variations through systematic $\mathcal{B}$-$\mathcal{G}$ pairings. Each group of selected tools will generate 2-3 scenarios. To ensure data diversity, if the similarity of a newly generated scenario exceeds a threshold $\epsilon$ compared to already synthesized scenarios, it will be excluded. Through this process, we can obtain a large number of synthetic scenarios, where the selected tools will serve as the "gold tools" for completing the corresponding virtual scenario, which will later be used for evaluation purposes.
\begin{align}
d((\mathcal{B}_i, \mathcal{G}_i), (\mathcal{B}_j, \mathcal{G}_j)) < \epsilon.
\end{align}

\subsection{Action Knowledge Exploration}
\paragraph{Initialization}
The root node is initialized with predefined Action Knowledge, which serves as the foundation for task-solving logic. 
During the MCTS process, the UCB algorithm is used to select nodes, effectively balancing exploration and exploitation by choosing the node with the highest upper confidence limit.

\paragraph{Expansion}
Upon selecting node $N_i$ as the candidate, an optimization process is initiated that retraces $N_i$ to obtain insights from previous optimization experience $\mathcal{E}$. 
Each of these past optimization experiences $\mathcal{E}$ is composed of three elements: the pre-optimization score $S_{before}$, the post-optimization score $S_{after}$, and the modification $\mathcal{M}$ of the optimization actions taken.
\begin{align}
\mathcal{E} =  \{ (S_{\text{before}}^i, S_{\text{after}}^i, \mathcal{M}^i) \mid N_i \in \text{Path}(N, N_0) \}
\end{align}

Based on the optimization experiences and exploration trajectories $Tra$ from the past, the LLM-based agent $\pi$ will analyze the discrepancies between the existing Action Knowledge and the environment. It will then optimize these to produce an updated version of the Action Knowledge.
\begin{align}
    \mathcal{AK}_{new} = \pi_\theta(\mathcal{AK}_{old}, \mathcal{E}, Tra)
\end{align}

\paragraph{Feedback Collection}
Once equipped with an optimized $\mathcal{AK}$, the agent $\pi$ can explore the environment to perform tasks. For each individual task $T$, the agent interacts with the environment to receive feedback with the trajectory $Tra_i$ and the final reward scores $S_i$. The score is related to the evaluation method of the task.
\begin{align}
    {Tra}_i,S_i = Env(\mathcal{AK},\pi)
\end{align}

\section{Experiment}

\subsection{Experiment Setup}
\paragraph{Datasets and Baselines}
To demonstrate the efficiency of our approach in optimizing action knowledge, we selected two datasets: ToolBench \cite{DBLP:conf/iclr/QinLYZYLLCTQZHT24} and HotpotQA \cite{DBLP:conf/emnlp/Yang0ZBCSM18}, each offering unique challenges for a comprehensive evaluation.
Following \citet{DBLP:journals/corr/abs-2410-08197}, several strong methods are selected as our baselines, including ReAct \cite{DBLP:conf/iclr/YaoZYDSN023}, Self-Refine \cite{DBLP:conf/nips/MadaanTGHGW0DPY23}, Easy-Tool \cite{DBLP:journals/corr/abs-2401-06201}, and DRAFT \cite{DBLP:journals/corr/abs-2410-08197}. See detailed setting and evaluation in Appendix~\ref{app: setting}.







\begin{table}[h]
\small
\centering
\resizebox{\columnwidth}{!}{%
\begin{tabular}{@{}c|l|cc|c}
\toprule
\multirow{2}*{\textbf{Model}}&\multirow{2}*{\textbf{Method}}& \multicolumn{2}{c|}{\textbf{ToolBench}} & 
\multicolumn{1}{c}{\multirow{2}*{\textbf{HotpotQA}}} \\
  & &\multicolumn{1}{c}{PASS} &\multicolumn{1}{c|}{WIN} &  \\
\midrule
\multirow{5}*{\textbf{GPT-4-turbo}}&

ReAct & 50.67 & 67.00 & 54.61  \\
&Self-Refine & 56.80 & \textbf{73.00} & 55.85 \\
&EasyTool & 51.67 & 68.00  & 58.19 \\
&DRAFT & 54.83 &  72.00  &  57.71 \\
&\cellcolor{mygray}{\textbf{Ours}} & \cellcolor{mygray}{\textbf{59.33}} & \cellcolor{mygray}{\textbf{73.00}} &\cellcolor{mygray}{\textbf{59.93}}\\
\midrule
\multirow{5}*{\textbf{Qwen-long}} &
ReAct & 48.30 & 71.00 & 52.00 \\
&Self-Refine & 53.70 & 77.00  &56.10\\
&EasyTool & 50.80 & 63.00  & 58.34 \\
&DRAFT &54.20 & 79.00 &53.23   \\
&\cellcolor{mygray}{\textbf{Ours}} &\cellcolor{mygray}{\textbf{57.20}} &\cellcolor{mygray}{\textbf{81.00}} & \cellcolor{mygray}{\textbf{59.91}}\\
\midrule
\multirow{5}*{\textbf{Qwen2-72B-Instruct}} &
ReAct & 49.43 & 55.00 & 50.21 \\
&Self-Refine & 54.33 & 65.00  & 52.59\\
&EasyTool & 52.97 & 58.00  & 54.94 \\
&DRAFT &56.43 & 69.00 & 57.57   \\
&\cellcolor{mygray}{\textbf{Ours}} &\cellcolor{mygray}{\textbf{58.52}} &\cellcolor{mygray}{\textbf{73.00}} & \cellcolor{mygray}{\textbf{58.70}}\\
\bottomrule
\end{tabular}
 }
\caption{\textbf{Main results of SynWorld compared to other baselines} on ToolBench and HotpotQA. The best results of each model are marked in \textbf{bold}. PASS means the pass rate and WIN means the win rate of the trajectory compared to GPT-3.5-turbo in the method of ReAct.}
\label{table: main}
\vspace{-5mm}
\end{table}






\subsection{Main Results}
\textbf{For the task ToolBench that requires the combined use of multiple tools}, as shown in Table~\ref{table: main}, our approach achieved a PASS score of 59.33 and a WIN score of 73.00, marking a significant improvement compared to other methods for iterative optimization, demonstrating the advantages of our method in terms of tool combination and task planning optimization.
\textbf{For the task HotpotQA that requires planning using a single tool and multi-step calls}, in the scenario where only a single tool is used but requires continuous multi-hop calls, our method has achieved state-of-the-art results. 
This indicates that we have not only aligned tool descriptions with the environment, but also succeeded in generating a generalizable planning workflow.

\subsection{Ablation Study} 
We observe that independently optimizing either the Workflow or the Tool Description using MCTS has its limitations in Table~\ref{tab:ablation}. 
We find that combining the optimization of both aspects leads to more effective results. 
An aligned Tool Description is beneficial for constructing a more reasonable Workflow, while a well-structured, general Workflow also enhances the exploration of tool usage.
We believe that this synergy arises during the iterative optimization process, where the improved workflow can help identify tool usage that is closer to the correct trajectory, serving as strong negative examples to further refine the tool description. 
Conversely, a superior tool description enables the model to generate workflows that are more aligned with the environment.
\begin{table}[h]
\small
    \centering
    \begin{tabular}{c|l|c}
        \toprule
        \textbf{Model} & \textbf{Method} & {\textbf{Pass Rate}} \\
        \midrule
                &  \cellcolor{mygray}{SynWorld} & \cellcolor{mygray}{59.33}  \\
        GPT-4-turbo & \quad w/o. Workflow &  56.33{\tiny\textcolor{red}{-3.00}} \\
        & \quad  w/o. Description & 53.16{\tiny\textcolor{red}{-6.17}}  \\

        \midrule
                & \cellcolor{mygray}{SynWorld} & \cellcolor{mygray}{57.20}  \\
        Qwen-long & \quad w/o. Workflow & 57.00{\tiny\textcolor{red}{-0.20}}  \\
        & \quad  w/o. Description & 53.83{\tiny\textcolor{red}{-3.37}} \\

        \bottomrule
    \end{tabular}
    \caption{Ablation experiment results}
    \vspace{-3mm}
    \label{tab:ablation}
    
\end{table}
\subsection{Futher Analysis}

\begin{figure}[tb]
    \centering
    \includegraphics[width=\linewidth]{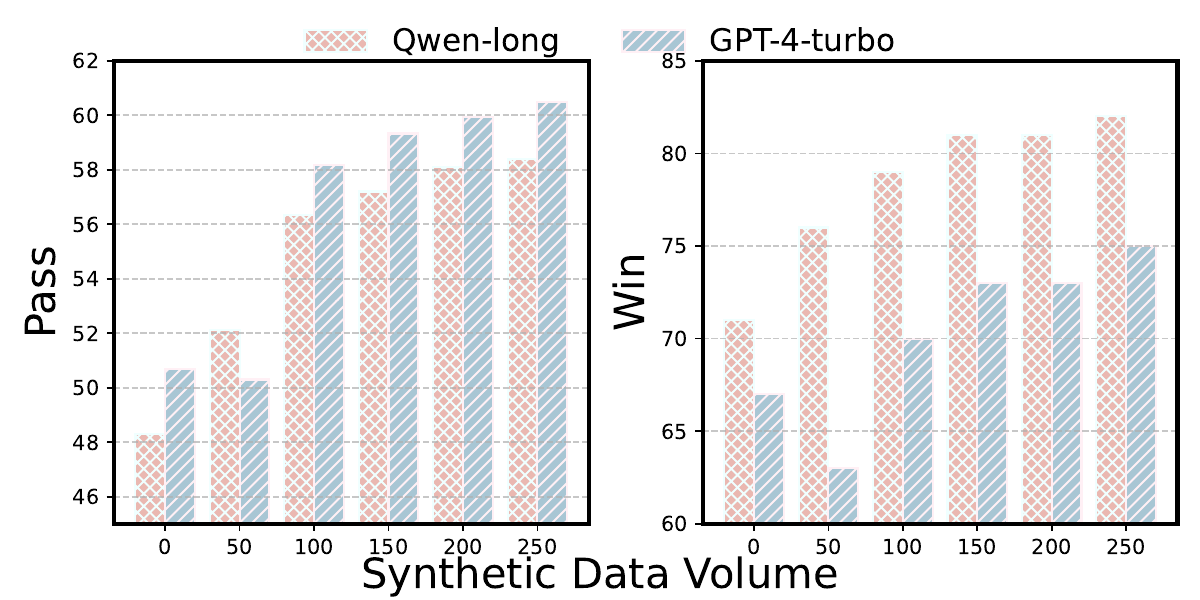}
    \caption{The variation in the pass rate of agents on the ToolBench in relation to the number of exploration scenarios.}
    \label{fig:sys_num}
\end{figure}

\paragraph{More simulated data enable precise virtual scenario synthesis, optimized action knowledge, and ultimately improved agent performance.} 

In our experiments, we explore action knowledge by synthesizing a varying number of virtual scenarios. As shown in Figure~\ref{fig:sys_num}, we find that as the number of scenarios synthesized increases, the performance of the Agent shows a corresponding upward trend. Specifically, within the range of 0 to 100 scenarios, the model's performance continues to improve with the increase in the scenarios, indicating that action knowledge is indeed learnable.
Although the rate of performance improvement slows down as the number of scenarios increases, the model's performance remains on an upward trajectory. 
This phenomenon suggests that the process of learning action knowledge in the context of synthesized scenarios exhibits scalability.

\paragraph{Virtual scenario policies can be generalized to unseen environments and improved with iterations.}

\begin{figure}[tb]
    \centering
    \includegraphics[width=0.9\linewidth]{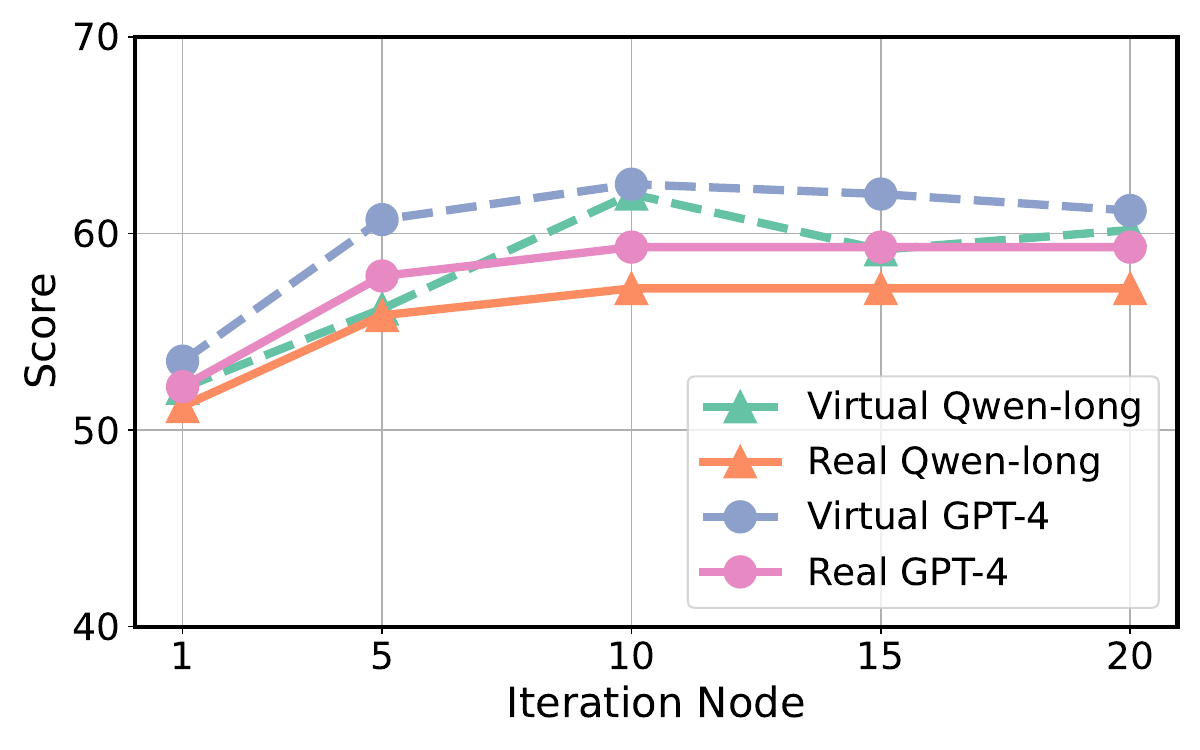}
    \caption{Changes in ToolBench pass rates in virtual and real-world scenarios with the number of iterative optimizations performed in the virtual environment.}
    \label{fig:exper2}
\end{figure}

By analyzing the relationship between action knowledge iterations and pass rates on Toolbench in both virtual and real environments, we find that the action knowledge gained in the virtual setting is generalizable and effective in real-world applications.
Performance trends in both environments are similar in Figure~\ref{fig:exper2}.
We observe a consistent upward trend in scores, particularly between 0 and 10 iterations, indicating that action knowledge can be optimized through environmental feedback. However, as iterations increase, the gains diminish, and we note slight declines in performance at times. This phenomenon is likely due to the limitations of exploring a fixed number of scenarios, where further iterations have less impact, and increasing complexity can hinder understanding.

\section{Conclusion}

In this paper, we propose SynWorld, a novel framework that synthesizes scenes that require multiple action steps and enhances agent action optimization through exploration in the synthetic virtual scenario.
By systematically exploring diverse synthetic scenarios, our model achieves precise alignment between action descriptions and environmental contexts while identifying task-specific workflows suitable for tasks.

\section*{Limitations}
We initially conduct empirical validation on two benchmarks: Toolbench (involving multi-tool calling scenarios) and HotpotQA (requiring multi-step action execution). While these demonstrate our method's effectiveness, broader validation across diverse real-world applications remains valuable. Promising candidates include web-based search tasks, simulated environments and so on.

Our approach currently incurs non-trivial computational overhead due to the token-intensive virtual scenario synthesis process. The exploration phase further compounds this by exhaustively enumerating all possible scenarios. Future research should prioritize optimizing token efficiency through 1) developing more economical synthesis mechanisms for high-quality virtual scenarios and 2) establishing effective filtering criteria to identify the most pedagogically valuable scenarios.

The current action knowledge representation employs a purely text-based format. This presents opportunities to investigate alternative structured representations that could enhance reasoning capabilities, such as tabular organization of action parameters or executable code snippets encapsulating procedural knowledge.

\section*{Acknowledgments}
This work was supported by the National Natural Science Foundation of China (No. 62206246, No. NSFCU23B2055, No. NSFCU19B2027), the Fundamental Research Funds for the Central Universities (226-2023-00138), Yongjiang Talent Introduction Programme (2021A-156-G), CIPSC-SMP-Zhipu Large Model Cross-Disciplinary Fund, Ningbo Natural Science Foundation (2024J020), Information Technology Center and State Key Lab of CAD\&CG, Zhejiang University. 
We gratefully acknowledge the support of Zhejiang University Education Foundation Qizhen Scholar Foundation.

\bibliography{custom}

\clearpage
\appendix

\section{Related Works}

\subsection{Agent Planning}

Recent studies has shown that in the realm of complex task-solving~\cite{DBLP:conf/emnlp/Ouyang023,sun2023adaplanner,liu2024tool,liu2025advanceschallengesfoundationagents}, the capacity for planning and refinement within large models has become increasingly pivotal. It has marked a transition from early Methods like CoT\cite{wei2022chain}, Plan and Solve~\cite{wang2023plan}, which tackle tasks sequentially, to the sophisticated agentic workflows of today, where model planning is instrumental in addressing a myriad of complex tasks, including question answering (QA)~\cite{mavi2022survey}, embodied interaction~\cite{yao2022webshop}, tool invocation~\cite{masterman2024landscape}, and long-form text generation~\cite{jiang2024longrag}. 

However, initial planning efforts are fraught with deficiencies due to the complexity of environments. When faced with unfamiliar environments, relying solely on human-written task descriptions without interaction with the environment often leads to plans that are misaligned with the actual tasks, or plans that seem reasonable but fail during execution due to a lack of accurate action knowledge. Consequently, there has been a surge in research~\cite{song2023llm,DBLP:journals/corr/abs-2302-01560} focused on refining plans and workflows. These efforts typically leverage direct environmental feedback or design a reward score for end-to-end plan correction, but they often lack a medium for the intermediate processes, which obscures the transparency of the plan refinement process. Moreover, the refinement of plans and the collection of feedback are usually linear and iterative~\cite{DBLP:journals/corr/abs-2410-08197}, resulting in low efficiency and a lack of diversity.

\subsection{Knowledge-Augmented Agents}
LLMs, as agents interacting with specific environments, often need to provide action signals to these environments~\cite{zhou2023agents,zhou2024symbolic,durante2024agent}. These action signals can be either restricted or open actions related to the environment~\cite{wang2024survey,li2024personal,hu2024survey,Zhangxilin}. For instance, they might involve specific movements in embodied task scenarios or the use of various tools like search or OCR in tool invocations. By incorporating these actions, on one hand, the Agent gains the ability to interact with the environment, allowing LLMs to transcend mere textual output~\cite{DBLP:conf/iclr/ShridharYCBTH21,DBLP:conf/emnlp/WangJCA22,SuifengZhao}. On the other hand, these external actions endow the Agent with a capability similar to humans using tools, compensating for the inherent limitations of LLMs, such as search tools that can alleviate issues of knowledge hallucination or obsolescence in LLMs~\cite{singh2024llm,zhao2024let,shen2024small,chen2023large}. 

Current methods for learning action knowledge are mainly divided into two categories: one involves creating a large amount of synthetic data to construct trajectories for executing actions to train the model~\cite{qiao2024autoact,huang2024understanding,zhu2024knowagent}, which is costly and has poor generalizability across different tasks; the other relies on prompt engineering~\cite{raman2022planning}, placing explicit action knowledge about how to plan to execute actions within the prompt, and then using ICL (In-Context Learning) methods to enable the model to learn to invoke these actions. While convenient, these methods can be inaccurate as the artificially constructed planning knowledge may not accurately reflect the true state of the environment, leading to potential biases.

\section{Setting}
\label{app: setting}
\subsection{Datasets}
\paragraph{ToolBench} contains tasks using over 16,000 RapidAPI tools. It assesses a model's ability to plan and execute complex workflows.
\paragraph{HotpotQA} is a multi-hop QA dataset with questions requiring multiple steps to answer. We employ Goolge Search as the search engine in the experiment.

\subsection{Evaluation}
ToolBench: Evaluated using pass rate and win rate. We record planning steps and tool invocations, then submit the trajectory for assessment. Win rate is compared to React's performance.
HotpotQA: Evaluated using F1 score, comparing model answers to gold answers (reward 0–1).
These datasets and metrics allow us to rigorously validate our approach across varied contexts.

\subsection{Baselines}
Several strong methods are selected as our baselines, including: ReAct \cite{DBLP:conf/iclr/YaoZYDSN023} which interacts with environment to reason the next step, Self-Refine \cite{DBLP:conf/nips/MadaanTGHGW0DPY23} which uses the feedback from environment to refine the origin prompt, Easy-Tool \cite{DBLP:journals/corr/abs-2401-06201} which uses llm firstly to refine the tool description and then break down the tasks to complete them, and DRAFT \cite{DBLP:journals/corr/abs-2410-08197} to synthesize tasks on a single tool for exploration to learn how to use the tool.

\subsection{Experiment Setup}
The backend model used in our experiments is Qwen-Long-0916, while the version of GPT-4 is 0613. The token usage in our method is approximately 6-8 million tokens. We configured the width of MCTS to 3 and set the similarity threshold to 0.6. After balancing effectiveness and cost, we synthesized 200 scenes and conducted 15 iterations on them during the experiment.

\section{Prompt Template}
See in Table~\ref{tab:prompt_tool},~\ref{tab:prompt_workflow}
\begin{table*}[t!]

    \centering
    \renewcommand\arraystretch{1}
    \scalebox{1.}{
    \begin{tabular}{p{14cm}}
    \toprule
    \large{\textbf{Prompt for Tool Description in Action knowledge }}\\
    \hline
Analyze the following tool execution trajectories to improve tool interface documentation.\\
For all trajectories:\\
1. Identify functional mismatches between original description and actual usage patterns\\
2. Detect parameter inefficiencies (missing/underutilized fields)\\
3. Extract implicit requirements from error patterns\\
4. Generate enhanced documentation with:\\
    Clear input specifications (required vs optional)\\
    Contextual usage guidelines\\
    Error prevention tips\\
    Response format expectations\\
Here is an \textbf{example}.\\
Now it's your turn to analyze the following tool execution trajectories to improve tool interface documentation.\\
tool\_name: \textbf{tool\_name}\\
original\_description: \textbf{original\_description}\\
trajectory: \textbf{trajectory}\\
Please provide your Optimize Description for the tool. Just modify the description part and do not change the parameters description.\\
Make Sure your description is clear and concise.\\
     \bottomrule
    \end{tabular}
    }
    \caption{Prompt used for tool document refinement.}
    \label{tab:prompt_tool}

\end{table*}
\begin{table*}[t!]
    \centering
    \renewcommand\arraystretch{1}
    \scalebox{1.}{
    \begin{tabular}{p{14cm}}
    \toprule
    \large{\textbf{Prompt for Workflow in Action knowledge }}\\
    \hline
Analyze the provided interaction trajectory and existing workflow steps to derive a generalized, reusable workflow for similar tool calling tasks.\\
1. Analyzing error patterns (authentication gaps, deprecated endpoints) and tool dependencies from interaction histories.\\
2. Extracting implicit requirements (authentication, sorting logic) and mandatory parameters from error responses.\\
3. Structuring a generic workflow with authentication validation, parameter checks, state management between API calls, and error fallbacks.\\
Here is an \textbf{example}.\\
Now it's your turn.\\
Existing Workflow: \textbf{workflow}\\
Trajectory: \textbf{trajectory}\\
Please provide your Optimize Workflow for the task. And make sure your workflow is clear and concise and no longer than 200 words.\\
     \bottomrule
    \end{tabular}
}
    \caption{Prompt used for workflow generation.}
    \label{tab:prompt_workflow}

\end{table*}

\section{Algorithm}
See in Algorithmic~\ref{alg:mcts}
\begin{algorithm*}
\caption{Monte Carlo Tree Search (MCTS) for Action Knowledge Optimization}
\label{alg:mcts}
\begin{algorithmic}
\Function{MCTS}{$root\_node$}
    \State $Iteration \gets 0$
    \While{$Iteration < max\_iteration$}
        \Comment{Step 1: Selection with UCB algorithm and $num\_child < 3$}
        \State $leaf\_node \gets \Call{select\_node}{root\_node}$
        
        \Comment{Step 2: Expansion}
        \State $new\_node \gets \Call{expand}{leaf\_node}$
        
        \Comment{Step 3: Simulation}
        \State $simulation\_result \gets \Call{simulate}{new\_node}$
        
        \Comment{Step 4: Backpropagation}
        \State $\Call{backpropagate}{new\_node, simulation\_result}$
        
        \State $Iteration \gets Iteration + 1$
    \EndWhile
\EndFunction

\Function{select\_node}{$node$}
    \While{$node.is\_fully\_expanded()$}
        \State $node \gets \Call{choose\_best\_child}{node, exploration\_parameter}$
    \EndWhile
    \State \Return $node$
\EndFunction

\Function{expand}{$node$}
    \State $optimization \gets \Call{choose\_untried\_optimization}{node}$
    \State $new\_node \gets \Call{apply\_optimization}{node, optimization}$
    \State $\Call{add\_child}{node, new\_node}$
    \State \Return $new\_node$
\EndFunction

\Function{simulate}{$node$}
    \State $optimized\_score \gets \Call{calculate\_score}{node.current\_action\_knowledge}$
    \State $reward \gets optimized\_score - father\_score$
    \State \Return $reward$
\EndFunction

\Function{backpropagate}{$node, result$}
    \While{$node \neq \textbf{None}$}
        \State $\Call{update\_statistics}{node, result}$
        \State $node \gets node.parent$
    \EndWhile
\EndFunction

\Function{calculate\_score}{$action\_knowledge$}
    \State \Return $\Call{evaluate}{action\_knowledge}$
\EndFunction
\end{algorithmic}
\end{algorithm*}

\appendix

\end{document}